# Revolutionizing Retail Analytics: Advancing Inventory and Customer Insight with AI

Ahmed Hossam
*Arab Academy for Science, Technology, and Maritime*
Alexandria, Egypt
a.h.khalil1@student.aast.edu

Ahmed Ramadan
*Arab Academy for Science, Technology, and Maritime*
Alexandria, Egypt
a.r.abdelrazik@student.aast.edu

Mina Magdy
*Arab Academy for Science, Technology, and Maritime*
Alexandria, Egypt
m.m.abdelkodos@student.aast.edu

Raneem Abdelwahab
*Arab Academy for Science, Technology, and Maritime*
Alexandria, Egypt
r.a.mohamed2@student.aast.edu

Salma Ashraf
*Arab Academy for Science, Technology, and Maritime*
Alexandria, Egypt
s.a.shaaban@student.aast.edu

Zeina Mohamed
*Arab Academy for Science, Technology, and Maritime*
Alexandria, Egypt
z.m.ibrahim1@student.aast.edu

*Abstract*—In response to the significant challenges facing the retail sector, including inefficient queue management, poor demand forecasting, and ineffective marketing, this paper introduces an innovative approach utilizing cutting-edge machine learning technologies. We aim to create an advanced smart retail analytics system (SRAS), leveraging these technologies to enhance retail efficiency and customer engagement. To enhance customer tracking capabilities, a new hybrid architecture is proposed integrating several predictive models. In the first stage of the proposed hybrid architecture for customer tracking, we fine-tuned the YOLOV8 algorithm using a diverse set of parameters, achieving exceptional results across various performance metrics. This fine-tuning process utilized actual surveillance footage from retail environments, ensuring its practical applicability. In the second stage, we explored integrating two sophisticated object-tracking models, BOT-SORT and ByteTrack, with the labels detected by YOLOV8. This integration is crucial for tracing customer paths within stores, which facilitates the creation of accurate visitor counts and heat maps. These insights are invaluable for understanding consumer behavior and improving store operations. To optimize inventory management, we delved into various predictive models, optimizing and contrasting their performance against complex retail data patterns. The GRU model, with its ability to interpret time-series data with long-range temporal dependencies, consistently surpassed other models like Linear Regression, showing 2.873% and 29.31% improvements in R2-score and mAPE, respectively.

## I. INTRODUCTION

The swift progress in Machine Learning, Computer Vision, and increased computational capabilities has opened up opportunities for the widespread integration and enhancement of machine learning across various industries [1]. In the retail industry, retailers face difficulties in making timely adjustments to optimize in-store operations and improve the overall shopping experience for customers. Challenges, such as inaccurate operational processes and imbalanced inventory levels, truly hindered informed decision-making. This, in turn, causes a loss in sales and customers. Consequently, huge efforts have to be implemented to develop strategic plans that can meet customer demands and capitalize on market trends [2].

In smart retail, accurate demand prediction is crucial for managing inventory levels effectively. By employing time series analysis [3], our system aims to anticipate the most frequently purchased products each season. This approach addresses the critical problems of stockouts (inventory shortages) and overstocking (excessive inventory), which can lead to lost sales, increased holding costs, and inefficient use of storage space. By enhancing demand forecasting, we can mitigate these risks, ensuring that retailers avoid both stockouts and overstocking, thereby optimizing inventory management and reducing associated costs.

To enhance in-store efficiency and customer satisfaction, the system will implement advanced customer tracking [4]. This approach addresses challenges like optimizing customer flow to reduce checkout queues and minimize wait times and congestion. It also plays a crucial role in staff allocation, enabling retailers to align human resources with customer demand effectively and avoid overstaffing or understaffing. Additionally, by incorporating machine learning techniques, the system aims to boost productivity through improved operational efficiency, optimizing resource utilization, and streamlining manual tasks within the retail environment.

Overall, SRAS aims to enhance the retail experience, improve operational efficiency, and enable data-driven decision-making for retailers. The structure of the paper is organized as follows: Section 2 discusses the Research Methodology, detailing the approaches and models used in this study. Section 3 presents the Experiments and Results, where we delve into the practical applications of our methodologies and the outcomes of our experiments.

## II. THE PROPOSED MODEL

In this section we introduce the applied methodologies, including object detection and tracking, and demand forecasting, evaluating their applicability and performance in enhancing retail industry operations and enriching the customer experience. Shown in Figure 1 is the architecture for our proposed model.

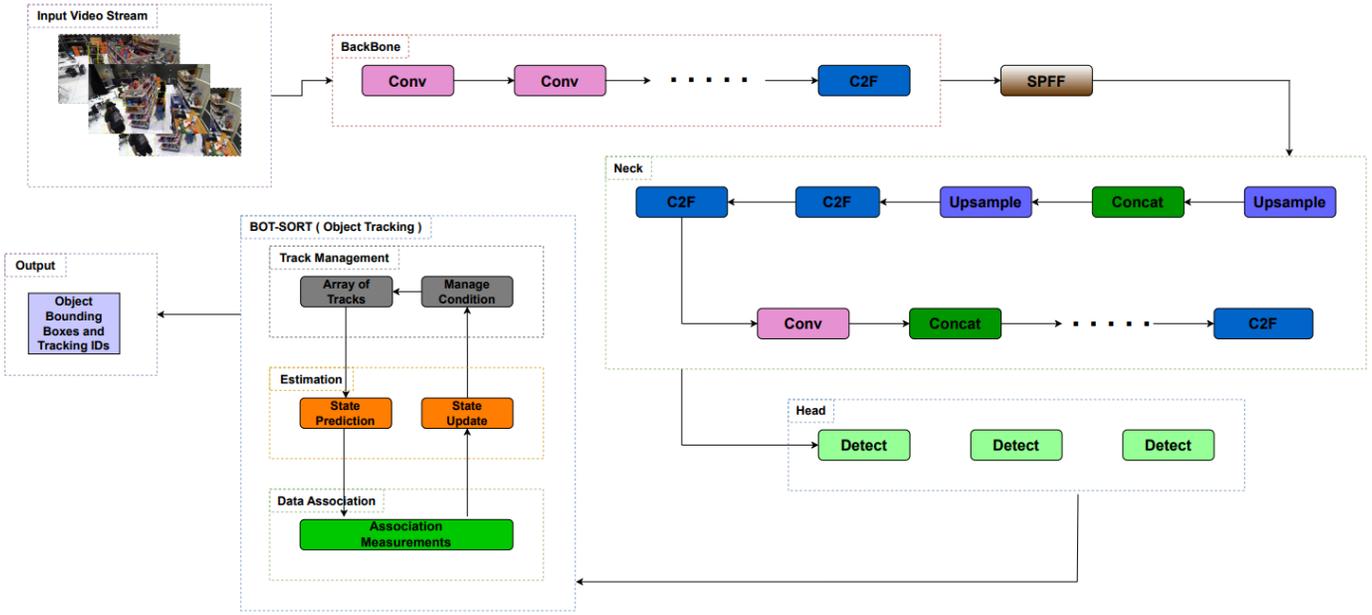

Figure 1

PROPOSED MODEL ARCHITECTURE, YOLO-V8 INTEGRATING WITH BOT-SORT

### A. Proposed Model

Our proposed model integrates two essential components, the first part employs YOLO-V8 for accurate and real-time person detection, while the second part incorporates the tracking capabilities using the BOT-SORT model. This integration ensures precise identification and continuous tracking of persons, making our model a robust solution for our demands.

The fundamental architecture of the YOLO model consists of three distinct components [5] as shown in Figure 1, that contribute to its robustness in object detection, (1) **The backbone:** considered the foundational element, it extracts features from any image input. (2) **The neck:** enhances the feature maps by combining information from diverse scales. (3) **The head:** it's the last layer of the YOLO architecture, it undertakes the crucial task of predicting bounding boxes.

The BOT-SORT algorithm in the proposed model uses a structured method for tracking. 'Track Management' oversees track statuses, while 'Estimation' employs Kalman Filter techniques for predicting and updating object states. 'Data Association' connects detections across frames, ensuring stable tracking in dynamic retail settings, and enhancing overall system performance [6].

### B. Object Detection

It is the initial step where the computer vision system identifies and localizes objects of interest within individual frames of a video or an image. Detection involves multiple steps: classification, localization, and detection, as shown in Figure 2. Classification assigns labels to the entire image, while localization determines the position of the identified object within the image, detection integrates classification and localization to identify and locate multiple objects within an image [7].

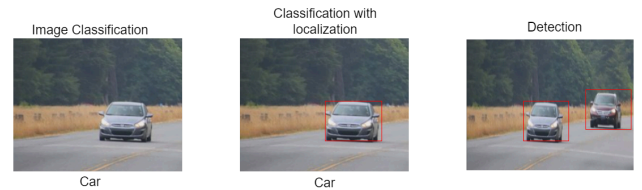

Figure 2

DIFFERENCE BETWEEN CLASSIFICATION, LOCALISATION, AND DETECTION

**YOLO** algorithm utilizes a single neural network for predicting the class probabilities and bounding boxes of objects in an image, as illustrated in Figure 3. Unlike Faster R-CNN architecture, which employs a multi-stage detection process [8], YOLO's approach processes the entire image in a single forward pass, enabling real-time object detection with impressive accuracy [5]. YOLO's real-time detection and accuracy align with the demands of our application making it the preferred choice over Faster R-CNN [9].

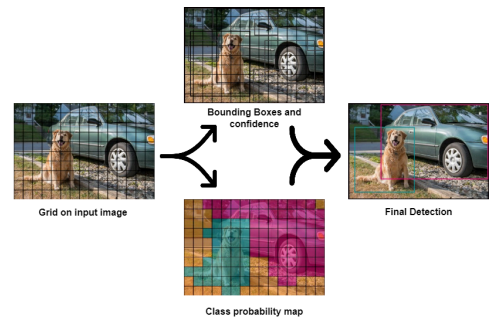

Figure 3

YOLO FLOW

In scenarios prioritizing individual detection, YOLO excels with its optimized architecture, identifying persons in diverse settings, Whether detecting a single person or multiple persons as shown in Figure 4.

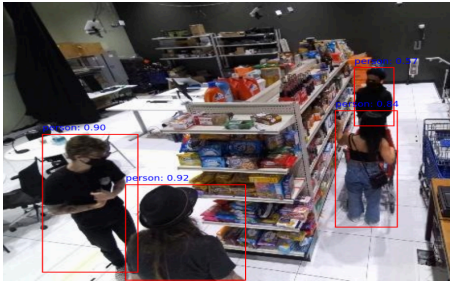

Figure 4

MULTIPLE PEOPLE DETECTION USING YOLO-V8

*Evolution of YOLO Models: YOLOv1 to YOLOv8*

The architectural evolution and performance of the YOLO models are concisely illustrated in Table I. It details the various YOLO versions alongside their respective frameworks, backbones, and improvements, highlighting the advancements up to YOLOv8. Showcasing the trade-offs between different YOLO variants, with YOLOv8 achieving an optimal balance.

TABLE I

COMPARISON OF DIFFERENT YOLO VERSIONS[14]

| YOLO Version | Framework | Backbone | Improvement/Challenges |
|---|---|---|---|
| V1 | Darknet | Darknet-24 | Faced challenges with small objects and spatial constraints |
| V2 and V3 | Darknet | Darknet-24 | Significantly improved small object detection |
| V4 | Darknet | CSPDarknet53 | Boosted accuracy and efficiency for real-time applications |
| V5 | PyTorch | Modified CSPv7 | Offered models of varying sizes for speed-accuracy trade-offs |
| V6 and V7 | PyTorch | CSPNet | Improved normalization and activation function |
| V8 | PyTorch | Advanced CSPNet | Exceptional balance of speed and accuracy. |

The YOLO-V8 model, equipped with 68.2 million parameters and 257.8 billion FLOPs, has proven to be outstanding. According to official YOLO documentation [5], it achieves an impressive 90% mean average precision on the COCO dataset. Additionally, the model operates with a low latency of just 3 milliseconds, making it highly efficient for real-time detection.

C. Object Tracking

*General Overview of Object Trackers*

Object tracking in the context of machine learning and computer vision is a technology that identifies and follows objects through a sequence of frames. This involves detecting an object in the first frame and then tracking its movement across frames. The primary challenge is to maintain the identity of the object despite changes in appearance, lighting, or occlusion. Object trackers use various cues such as appearance, motion, and sometimes additional features like shape to continuously predict the object's location in each frame. Sophisticated algorithms, such as Kalman filters or deep learning models, are often employed to analyze temporal and spatial information [4].

*Performance Evaluation*

This section examines the performance of (SOTA) object-tracking algorithms. The authors of BoT-SORT presented a side-by-side comparison of these methods on the MOT17 test set [6]. BoT-SORT stands out with the highest MOTA score of 80.6, highlighting its strength in inaccurate identity tracking with fewer errors. The algorithm's effectiveness is enhanced by its lower false positives and negatives, making it valuable in dynamic retail environments where tracking precision drives customer behavior insights and operational improvements. ByteTrack, while faster, shows a trade-off with slightly less accuracy. This evaluation stresses the need for a balanced tracker that aligns with application-specific demands, whether that be speed or accuracy.

*BOT-SORT Overview*

BOT-SORT is a multi-object tracking algorithm that shows remarkable results in crowded and complex scenes. It incorporates techniques like a modified Kalman Filter and Camera Motion Compensation (CMC) to improve tracking accuracy. The modified Kalman Filter is particularly effective in fitting bounding boxes more accurately to objects [6]. BOT-SORT also addresses the challenges posed by camera movement, especially in dynamic retail environments, making it highly suitable for our study.

*ByteTrack Overview*

ByteTrack is known for its innovative approach to multi-object tracking by associating every detection box. It handles scenarios where missing detections and low-scoring detections occur, often due to occlusions or motion blur. ByteTrack leverages information from previous frames to enhance video detection performance and uses a combination of high and low-score detection boxes for more comprehensive tracking[4].

D. Demand Forecasting

Demand prediction plays a crucial role in supply chains and builds the basis for business operations. This is because customer demand is the starting point for the decision-making process, aiming to reduce the gap between the forecasted and the current demand. Inaccurate demand forecasting will cause a huge loss in sales, as understocking leads to customer dissatisfaction, while overstocking leads to increased costs per inventory [10].

In this paper, the following models have been applied to determine which model will produce the most accurate prediction: (1) **Linear Regression**: It's a fundamental machine learning algorithm used to model the association between a reliant variable and one or several independent variables. (2) *XGBoost (Extreme Gradient Boosting):* is an ensemble learning algorithm based on decision trees. It employs gradient boosting to improve the performance of weak models progressively. Known for its accuracy and high efficiency, XGBoost has been used in a variety of applications in various domains including sales forecasting.

(3) **CNN (Convolutional Neural Network):** CNN is a deep learning model that specializes in processing structured grid-like data, such as time-series data. While CNNs are widely used in computer vision tasks, they can also be adapted for time-series analysis, including sales forecasting. In this study, CNN was employed to extract relevant features from the sales data and capture patterns that are indicative of future sales trends. (4) **LSTM (Long Short-Term Memory):** A variant of neural network that is suitable for capturing long-term dependencies in time-series analysis. LTSM is widely applied in NLP (Natural Language Processing) and word recognition but can also be used for sales forecasting. (5) **GRU (Gated Recurrent Unit):** Similar to LTSM, this type of recurrent neural network (RNN) excels at capturing long-term dependencies in time-series data. Unlike traditional RNNs, GRU incorporates gating mechanisms to selectively retain and update information. Although GRU is commonly used in NLP and speech recognition, it can also be effectively employed for sales forecasting. In this study, GRU was utilized to capture temporal patterns in sales data.

### III. EXPERIMENTS AND RESULTS

This section will illustrate various scenarios involving YOLO models and different demand forecasting models.

#### A. Object Detection, and Tracking

*1) Datasets:* The proposed model was evaluated using the MMPTRACK dataset [11] a large-scale benchmark dataset for people tracking. The statistics of the collected dataset are summarized in Table II. Our dataset is recorded with 15 frames per second(FPS) in five diverse environments.

TABLE II
STATISTICS OF MULTI-CAMERA MULTIPLE PEOPLE TRACKING (MMPTRACK) DATASETs

| Envs | Retail | Lobby | Industry | Cafe | Office | Total |
|---|---|---|---|---|---|---|
| # of cameras | 6 | 4 | 4 | 4 | 5 | 23 |
| Train (min) | 84 | 65 | 52 | 14 | 46 | 261 |
| Validation (min) | 43 | 32 | 31 | 28 | 19 | 153 |
| Test (min) | 45 | 32 | 32 | 31 | 22 | 162 |
| Total (min) | 172 | 129 | 115 | 73 | 87 | 576 |

Overall, about 9.6 hours of videos were collected, with over half a million annotations for each camera. The dataset is fully annotated with person bounding boxes and corresponding IDs. All videos are recorded with cameras placed at different angles, with a guarantee that all cameras' fields of view are connected (one camera has overlapped FoV with at least one of the other cameras). For each frame, Customer classes were annotated. Samples from the Dataset in Figure 5 show data collection from different cameras.

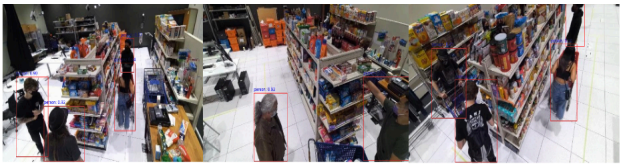

Figure 5
SAMPLES FROM THE DATASET FOR CUSTOMER DETECTION

*2) Evaluation Metrics:* To evaluate the proposed model, accuracy, recall, precision, and mAP are employed. The most important and commonly used metric in evaluating customer detection and detection is the **Mean Average Precision (mAP).** Mean average precision (mAP) is a quantitative measure used to assess the effectiveness of object detection models. The mean Average Precision (mAP) is computed by taking the average of the Average Precision (AP) values for each class in the model. Average Precision (AP) quantifies the model's ability to accurately identify items belonging to a specific class across various confidence criteria. Average Precision (AP) is calculated by graphing the precision-recall curve for each class and determining the area under the curve. mAP, or mean Average Precision, is a metric that evaluates the performance of a model by taking into account both the precision and recall of the detections. It provides a comprehensive measure of both the accuracy and completeness of the model.

*3) Experimental Results:* We applied YOLOV8-N, YOLOV8-S, YOLOV8-M, YOLOV8-L, and YOLOV8-X into the customer detection and tracking model to compare their performances and get the final results through the best performance model. The comparison results of the aforementioned models using evaluation metrics are shown in Table III.

TABLE III
EXPERIMENTAL RESULTS FOR PERSON DETECTION APPLIED USING MODELS

| Model Name | Optimizer | Epochs | Batch Size | Precision | Recall | MAP50 | MAP50-95 |
|---|---|---|---|---|---|---|---|
| yolov8n.pt | AdamW | 50 | 8 | 0.942 | 0.995 | 0.99 | 0.837 |
| yolov8s.pt | AdamW | 50 | 8 | 0.968 | 0.985 | 0.991 | 0.823 |
| yolov8m.pt | AdamW | 50 | 8 | 0.969 | 0.969 | 0.986 | 0.828 |
| yolov8l.pt | AdamW | 50 | 8 | 0.981 | 0.923 | 0.983 | 0.834 |
| yolov8x.pt | AdamW | 50 | 8 | 0.967 | 0.984 | 0.987 | 0.83 |

Due to the performance of the YOLOV8-X model, it has been used as the final model for customer detection and tracking. Table IV and Table V summarize the performance of the YOLOV8-X model based on different parameter combinations.

TABLE IV
EXPERIMENT RESULTS OF ALTERING OPTIMIZERS ON PERFORMANCE

| Model Name | Optimizer | Epochs | Batch Size | Precision | Recall | mAP50 |
|---|---|---|---|---|---|---|
| yolov8v.pt | Adam | 50 | 8 | 0.949 | 0.985 | 0.988 |
| yolov8v.pt | AdamW | 50 | 8 | 0.968 | 0.985 | 0.987 |
| yolov8v.pt | SGD | 50 | 8 | 0.97 | 0.998 | 0.987 |
| yolov8v.pt | RMSProp | 50 | 8 | 0.953 | 0.969 | 0.976 |

TABLE V
EXPERIMENT RESULTS WITH APPLYING DIFFERENT EPOCH SIZES USING ADAMW OPTIMIZER

| Model Name | Optimizer | Ephocs | Batch Size | Precision | Recall | mAP50 | Mao50-95 |
|---|---|---|---|---|---|---|---|
| yolov8x.pt | AdamW | 10 | 8 | 0.879 | 0.815 | 0.924 | 0.559 |
| yolov8x.pt | AdamW | 50 | 8 | 0.967 | 0.985 | 0.987 | 0.83 |
| yolov8x.pt | AdamW | 100 | 8 | 0.26 | 0.976 | 0.985 | 0.847 |
| yolov8x.pt | AdamW | 150 | 8 | 0.969 | 0.993 | 0.993 | 0.856 |

The YOLOV8-X model for customer detection and tracking has been fine-tuned on real footage from surveillance cameras located inside retail stores. It was tested on different occasions with different configurations. In one instance, a single layer was frozen, while in another, two layers were frozen.

Figure 6 below shows the performance measures, including Precision, Recall, mAP50, and Classification loss, which vary across epochs in the One-layer tuned model.

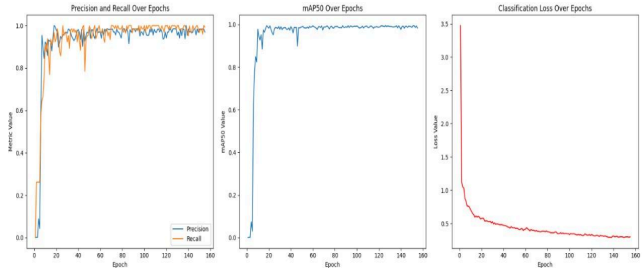

Figure 6
EFFECT OF TUNING ONE LAYER ON PERFORMANCE

Figure 7, on the other hand, illustrates how the same performance measures vary across epochs in the Two-layer tuned model.

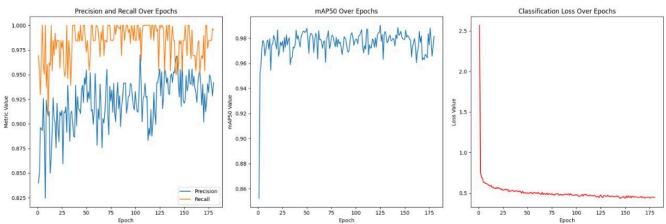

Figure 7
EFFECT OF TUNING TWO LAYERS ON PERFORMANCE

Based on the result as shown in Figure 6, the mAP value is considered outstanding, and the Precision and Recall measures are promising, affirming the re-trained YOLOV8-X model's high accuracy in detecting customers in retail stores. The decision to choose t for a 1-tune layered approach significantly influenced performance on the contrast of the 2-tuned layered approach, as demonstrated in Figures 6 and Figure 7. The 1-tune layered approach showed a successful pattern of performance as shown in Figure 6.

*Object Tracking Analysis and Choice*

Our experimental results of BOT-SORT and ByteTrack, two leading multi-object tracking technologies, were conducted against the backdrop of high-traffic retail environments. BOT-SORT edged out with a higher MOTA score, signifying its superior handling of false positives, false negatives, and identity switches—crucial metrics in densely populated scenarios. ByteTrack excelled in processing speed, achieving an average of 17 FPS, while BOT-SORT maintained a competitive 13 FPS. Despite ByteTrack's rapid frame rate, the precision of BOT-SORT in tracking multiple objects in tight spaces made it the preferable choice for our rigorous retail applications.

*B. Demand Forecasting Model*

*1) Dataset:* The Store Item Demand Forecasting Challenge dataset [12] is a comprehensive collection of data that aims to tackle the task of demand forecasting for various store items. This dataset is specifically designed to address the challenge of predicting future demand based on historical sales data. The dataset contains information about multiple stores and their corresponding items, along with the dates and sales figures, as shown in Table VI. Each record in the dataset represents a specific item sold at a particular store on a given date. The data spans a time, allowing for the analysis of trends and patterns over different time intervals.

TABLE VI
SAMPLE OF STORE ITEM DEMAND FORECASTING DATASET

| date | store | item | sales | month | week | day | daily_avg_sales | monthly_avg_sales |
|---|---|---|---|---|---|---|---|---|
| 10/2/2013 | 1 | 1 | 11 | 1 | 1 | 2 | 18.793103 | 13.709677 |
| 4/25/2013 | 5 | 2 | 28 | 4 | 17 | 25 | 44.007663 | 46.94 |
| 9/1/2016 | 5 | 6 | 53 | 9 | 35 | 1 | 44.386973 | 46.706667 |
| 9/8/2015 | 3 | 19 | 49 | 9 | 40 | 29 | 42.038314 | 47.38 |
| 10/8/2015 | 2 | 38 | 119 | 10 | 41 | 8 | 102.51341 | 102.806452 |
| 3/26/2015 | 9 | 43 | 56 | 3 | 13 | 26 | 53.83908 | 48.787097 |

*2) Evaluation Metrics:* To assess the different models' performance and accuracy, different metrics were applied, (1) **The Root Mean Square Error (RMSE)**: measures the average difference between a statistical model's predicted values and the actual values. Low RMSE values indicate that the model fits the data well and has more precise predictions. (2) **R-squared:** It represents the percentage of the variability in the dependent variable that is explained by the independent variables in the model. R² ranges from 0 to 1, where 1 indicates a perfect fit. Higher R² values indicate a better predictive performance of the model. (3) **Mean Absolute Percentage Error (MAPE)**: It calculates the average of the absolute percentage errors between the predicted and actual values. Lower MAPE values indicate higher accuracy of the forecasting method. (4) **Mean absolute error (MAE):** represents the difference between the original and predicted values extracted by averaging the absolute difference over the data set [13]. (5) **Percentage of improvement Rate:** calculates the absolute difference between the proposed model and the other models respectively, then divide the result by the initial score, and finally, multiplied by 100.

*3) Experimental Results:* In this study, Five Machine learning models were applied and their performances were measured, Linear Regression, XGBoost, CNN, LSTM, and GRU were evaluated and compared using retail store sales data. The purpose of this analysis is to determine which model performed the best in predicting the store's demand.

The result of this analysis showed that GRU outperformed the other models in terms of accuracy and efficiency. RMSE

and R-squared. It had achieved the lowest RMSE, and Highest R-squared compared to the other models. This indicates that GRU was able to capture more complex patterns and generate more accurate results.

Table VII, shown below, summarizes the comparison of the models' results.

TABLE VII
OUR EXPERIMENTAL RESULTS FOR DEMAND FORECASTING. ILLUSTRATING DIFFERENT PERFORMANCE METRICS ON DIFFERENT SCENARIOS

| Metric | Linear Regression | XGBoost | CNN | LSTM | GRU |
|---|---|---|---|---|---|
| RMSE | 9.325 | 7.998 | 8.002 | 8.359 | 7.983 |
| R2-score | 0.905 | 0.902 | 0.93 | 0.924 | 0.931 |
| MAPE | 0.174 | 0.135 | 0.124 | 0.127 | 0.123 |
| MAE | 7.273 | 6.239 | 6.164 | 6.382 | 6.133 |
| MSE | 86.956 | 62.734 | 64.023 | 69.869 | 63.725 |

Table VII demonstrates that the proposed GRU-based model exhibited exceptional performance, with improvement rates of 2.873%, 3.215%, 0.323%, and 0.756% in R2-score when compared to Linear Regression, XGBoost, CNN, and LSTM, respectively. Furthermore, in terms of mAPE, the GRU model showed significant improvement rates of 29.31%, 8.889%, 0.806%, and 3.149% relative to the same models, respectively.

Figures 8 and 9 display the training and validation losses for GRU, as well as the total amount of sales compared with the total predicted sales from 2017-10-01 to 2018-01-01.

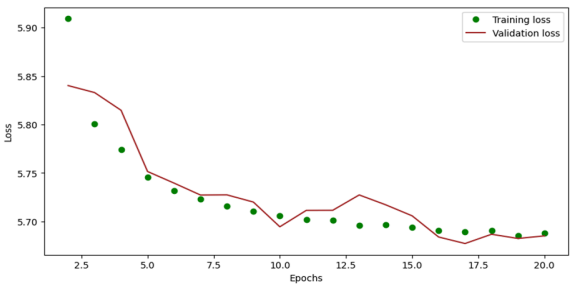

Figure 8
GRU LOSS ACROSS EPOCHS

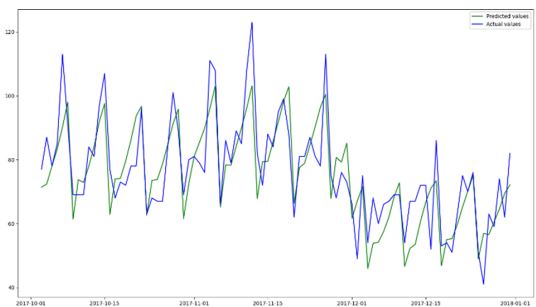

Figure 9
COMPARISON OF PREDICTED AND ACTUAL SALES

## IV. CONCLUSION

In conclusion, our study marks a significant step forward in retail optimization, combining the precision of YOLOV8 for customer detection with the advanced capabilities of BOT-SORT for detailed object tracking, all within a smart retail analytics system (SRAS). These technologies, alongside the GRU model's accurate demand prediction, form a comprehensive solution to modernize retail operations. The integration of these cutting-edge models promises a retail environment that is not only operationally efficient but also finely attuned to the changing needs of customers and businesses alike. Through our research, we've laid the foundation for an innovative, data-driven approach that can greatly enhance both the retailer and customer experience.